\documentclass[letterpaper]{article} 
\usepackage{aaai23}  
\usepackage{times}  
\usepackage{helvet}  
\usepackage{courier}  
\usepackage[hyphens]{url}  
\usepackage{graphicx} 
\usepackage{natbib}  
\usepackage{caption} 
\usepackage{algorithm}
\usepackage{algorithmic}
\usepackage{subfigure}
\usepackage{pifont}
\newcommand{\cmark}{\ding{51}}%
\newcommand{\xmark}{\ding{55}}%
\usepackage{multirow}
\usepackage{amsfonts}
\usepackage{amsmath}
\usepackage{booktabs}

\usepackage{newfloat}
\usepackage{listings}
\DeclareCaptionStyle{ruled}{labelfont=normalfont,labelsep=colon,strut=off} 
\floatstyle{ruled}
\newfloat{listing}{tb}{lst}{}
\floatname{listing}{Listing}
\title{HRDoc: Dataset and Baseline Method toward Hierarchical Reconstruction of Document Structures}
\author{
    Jiefeng Ma\textsuperscript{\rm 1},
    Jun Du\textsuperscript{\rm 1}\thanks{Corresponding author.},
    Pengfei Hu\textsuperscript{\rm 1},
    Zhenrong Zhang\textsuperscript{\rm 1},
    Jianshu Zhang\textsuperscript{\rm 2},
    Huihui Zhu\textsuperscript{\rm 2},
    Cong Liu\textsuperscript{\rm 2}
}
\affiliations{
    \textsuperscript{\rm 1}NERC-SLIP, University of Science and Technology of China\\
    \textsuperscript{\rm 2}iFLYTEK Research\\
    jfma@mail.ustc.edu.cn, jundu@ustc.edu.cn, \{hudeyouxiang, zzr666\}@mail.ustc.edu.cn,\\
    \{jszhang6, hhzhu2, congliu2\}@iflytek.com

}

\usepackage{bibentry}

\begin{document}
	
	\maketitle
	
	\begin{abstract}

		The problem of document structure reconstruction refers to converting digital or scanned documents into corresponding semantic structures.
		Most existing works mainly focus on splitting the boundary of each element in a single document page, neglecting the reconstruction of semantic structure in multi-page documents.
		This paper introduces hierarchical reconstruction of document structures as a novel task suitable for NLP and CV fields.
		To better evaluate the system performance on the new task, we built a large-scale dataset named HRDoc, which consists of 2,500 multi-page documents with nearly 2 million semantic units.
		Every document in HRDoc has line-level annotations including categories and relations obtained from rule-based extractors and human annotators.
		Moreover, we proposed an encoder-decoder-based hierarchical document structure parsing system (DSPS) to tackle this problem. 
		By adopting a multi-modal bidirectional encoder and a structure-aware GRU decoder with soft-mask operation, the DSPS model surpass the baseline method by a large margin.
		All scripts and datasets will be made publicly available at https://github.com/jfma-USTC/HRDoc.
	\end{abstract}
	
	\section{Introduction}
	
	\begin{figure}[t]
		\centering
		\subfigure[Multi-page documents]{
			\begin{minipage}[t]{0.5\linewidth}
				\centering
				\includegraphics[width=1.3in]{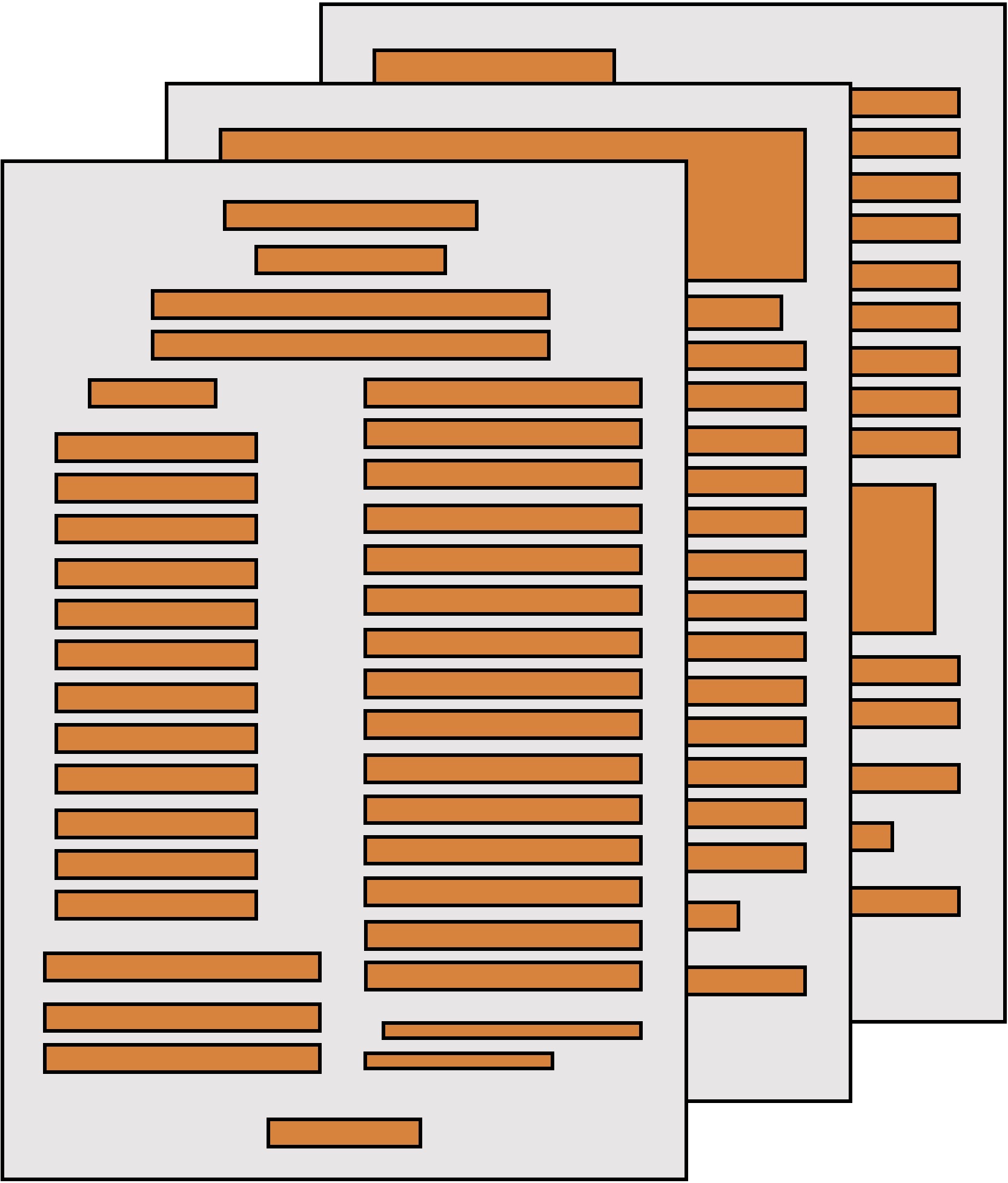}
				\label{fig:1a}
			\end{minipage}%
		}%
		\subfigure[Line-level classification]{
			\begin{minipage}[t]{0.5\linewidth}
				\centering
				\includegraphics[width=1.3in]{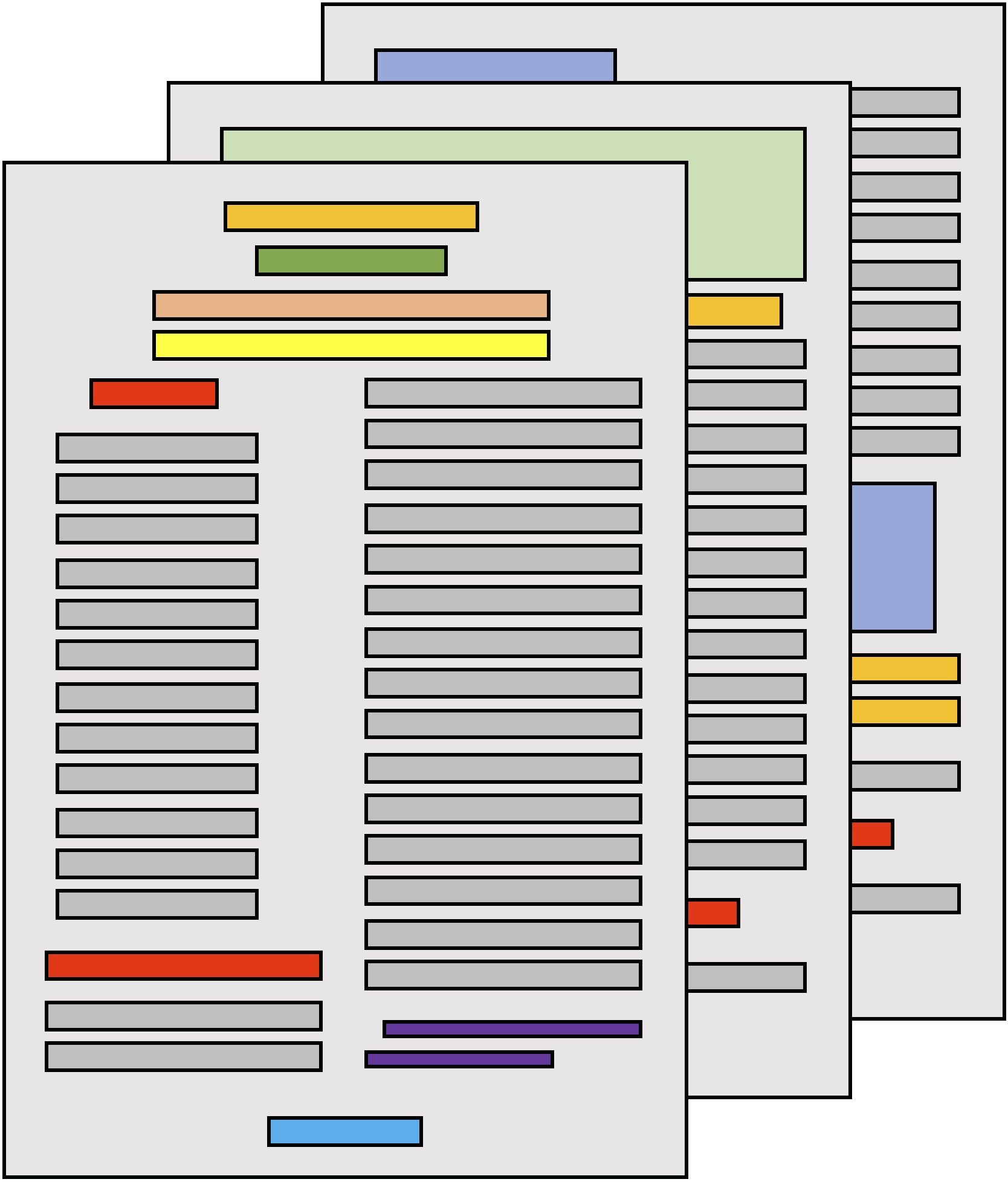}
				\label{fig:1b}
			\end{minipage}%
		}%
		\quad
		\subfigure[Hierarchical relationship]{
			\begin{minipage}[t]{0.5\linewidth}
				\centering
				\includegraphics[width=1.3in]{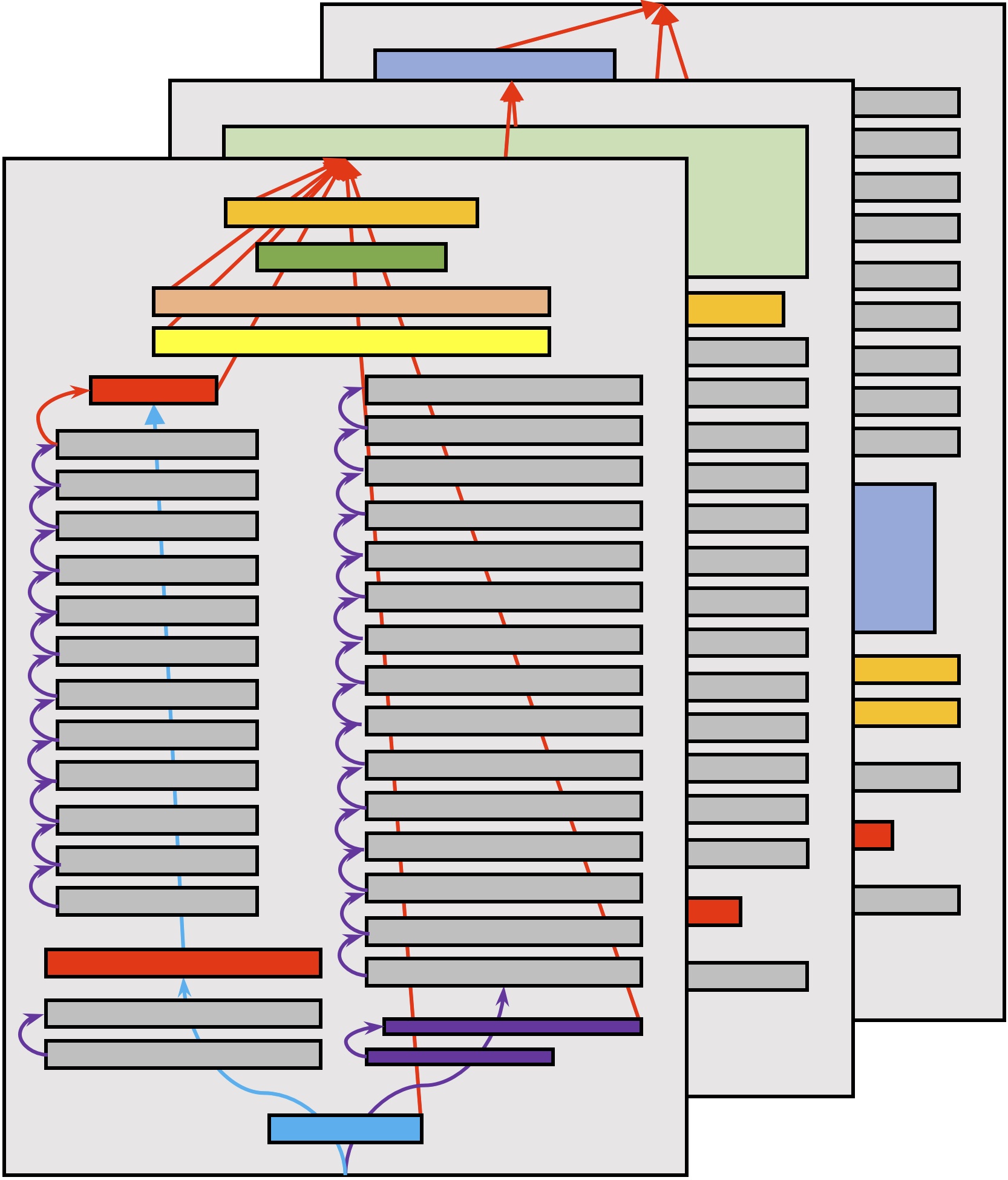}
				\label{fig:1c}
			\end{minipage}
		}%
		\subfigure[Semantic stucture]{
			\begin{minipage}[t]{0.5\linewidth}
				\centering
				\includegraphics[width=1.3in]{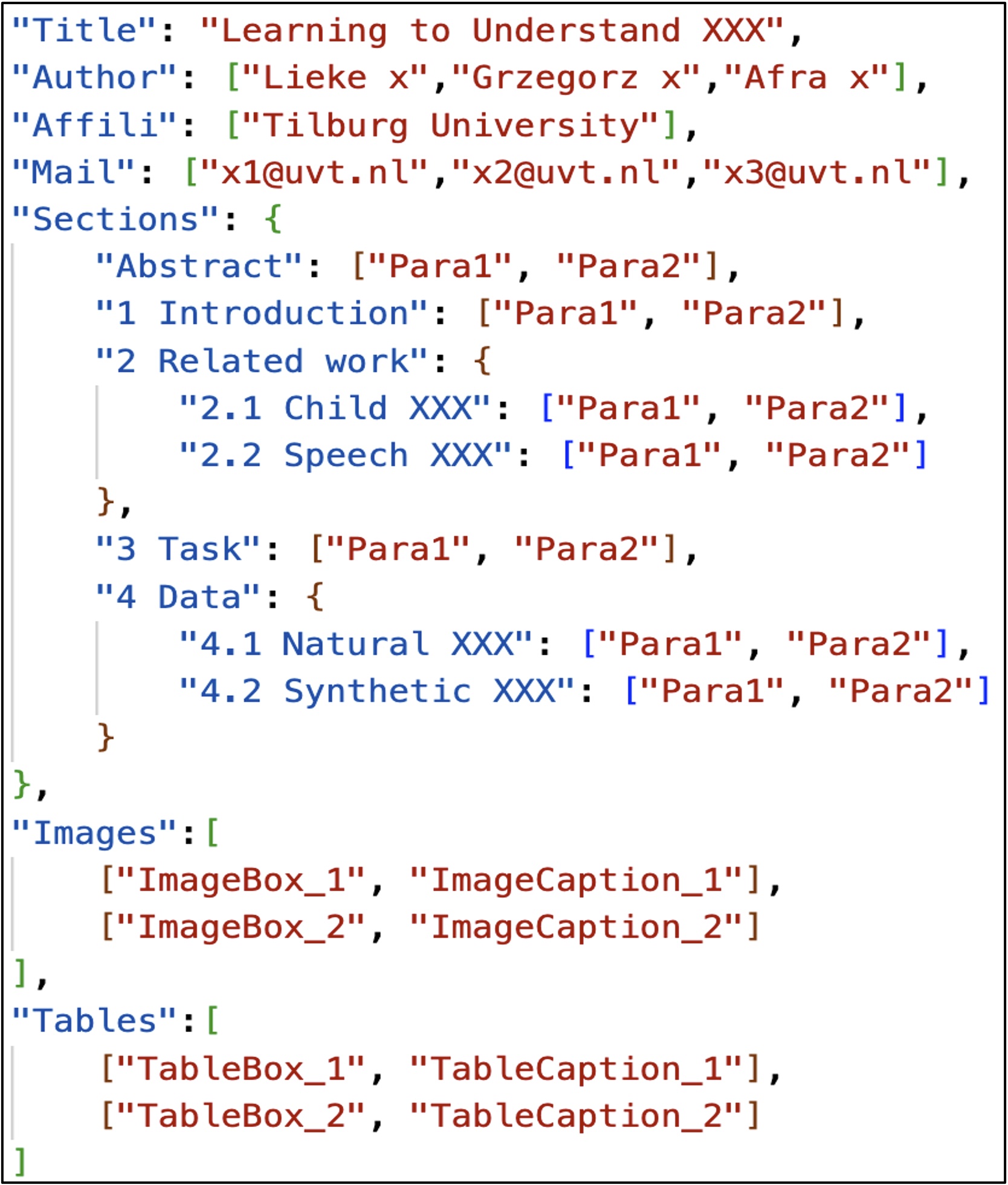}
			\end{minipage}
			\label{fig:1d}
		}%
		\centering
		\caption{The processing procedure of hierarchical document structure reconstruction task. }
		\label{fig:taskintro}
	\end{figure}
	
	With the prosperity of commercial activities in today's society, different documents are used to convey information, leading to a growing demand for automatic document processing \cite{cui2021document}.
	As an emerging application scenario in document processing, extracting or formalizing the structure of numerous documents is of great value to users. 
	For example, the document structure is required when converting a PDF file to an editable format such as Markdown.
	
	Current studies \cite{zhong_publaynet_2019, li_docbank_2020, pfitzmann_doclaynet_2022} focused predominantly on the document element detecting task and obtained considerable achievements in page-level layout analysis.
	Based on these works, some methods \cite{wang_docstruct_2020} and datasets \cite{rausch_docparser_2021} were proposed for extracting hierarchy structure in documents. Nonetheless, they neither achieved a satisfactory result nor provided a reasonable definition of the document structure reconstruction task.
	
	To move towards a more advanced understanding of hierarchical document structure reconstruction, we have created a new dataset named HRDoc, which consists of two parts according to the layout varieties. A multi-modal system is also proposed and demonstrates great superiority over the baseline method. 
	
	Figure \ref{fig:taskintro} depicts the general task definition and processing procedure. We define the document structure reconstruction task as converting one multi-page document (Figure \ref{fig:1a}) into its hierarchical semantic representation (Figure \ref{fig:1d}). 
	More specifically, we regard text lines, figures, tables, and equation areas as the essential elements of one document, which can be obtained through an off-the-shelf OCR engine \cite{Tesseract} or PDF extractor \cite{marinai2009metadata}. 
	After extracting these essential elements, we can use deep learning techniques to predict labels for each element unit (Figure \ref{fig:1b}) and the relation (Figure \ref{fig:1c}) between them.
	Combining the information above, we can recover the document's hierarchical semantic structure as shown in Figure \ref{fig:1d}.
	
	
	Our main contributions are listed as follows:
	\begin{itemize}
		\item We introduced the hierarchical reconstruction of document structures as a novel vision and language task.
		\item We built a novel hierarchical document structure reconstruction dataset HRDoc, 
		which is the first dataset that focuses on fine-grained and document-level structure reconstruction tasks over multi-page documents.
		\item We proposed an encoder-decoder-based hierarchical document structure parsing system (DSPS).
		With the multi-modal bidirectional extractor and the structure-aware GRU decoder with soft-mask operation, DSPS achieved considerable improvement over the baseline method.
	\end{itemize}
	The HRDoc dataset and source code of the proposed DSPS framework will be made publicly available here\footnote{https://github.com/jfma-USTC/HRDoc}.

	\section{Related Work}
	
	\subsection{Document Layout Analysis} 
	The most relevant task with document structure reconstruction in the document understanding field is layout analysis, which mainly focuses on document object detection and recognition of specific types such as figures and tables.
	
	Early works of document layout analysis relied on various heuristic rules or conventional image process methods to locate semantic regions in document images \cite{jaekyu_ha_recursive_1995, simon_fast_1997, kumar_text_2007}.
	In \citeyear{jaekyu_ha_recursive_1995}, \citeauthor{jaekyu_ha_recursive_1995} proposed a X-Y cut algorithm to determine the boundaries between each document component in a top-down manner. 
	\citeauthor{simon_fast_1997} regarded the connected components as the basic units in a binary document image, which are further combined into higher-level structures through heuristics methods and labeled according to structural features.
	\citeauthor{kumar_text_2007} used matched wavelets and MRF model to segment document images into text, background, and picture components at a pixel level.
	These approaches often used complicated hyper-parameters designed for specific document layouts, which are difficult to adapt to new document types. 
	
	Recent works take advantage of the wide-spreading machine learning techniques of object detection on natural scene images to address complex layout analysis problems \cite{he_multi-scale_2017, gao_deep_2017, yi_cnn_2017, li_page_2018}. 
	\citeauthor{he_multi-scale_2017} proposed a two-stage approach to detect tables and figures. In the first stage, the class label for each pixel is predicted using a multi-scale convolutional neural network. Then in the second stage, heuristic rules are applied to the pixel-wise class predictions to get the object boxes. \citeauthor{gao_deep_2017} utilized meta-data information from PDF files and detected formulas using a model that combines CNN and RNN. \citeauthor{yi_cnn_2017} redesigned the CNN architecture of common object detectors by considering the uniqueness of document objects. \citeauthor{li_page_2018} generated the primitive region proposals from each column region and clustered the primitive proposals to form a single object instance.
	
	These works mainly focus on splitting the boundary of each element in a single document page, neglecting the reconstruction of semantic structure in multi-page documents.

	\subsection{Document Structure Reconstruction}
	Document structure reconstruction means recovering the hierarchical semantic information of one document.
	Existing works on document structure reconstruction can be roughly divided into two categories.
	
	The first one focuses on reconstructing only part of the structure information in one document, such as the table of contents (ToC) \cite{wu2013table, tuarob_hybrid_2015, nguyen_enhancing_2017, bentabet-etal-2020-financial}.
	\citeauthor{tuarob_hybrid_2015} listed several position-related rules and used a random forest algorithm to determine whether one text line is a section name or not. With all sections detected, a set of regular expressions and rule-based strategies were used to recover the hierarchical structure of sections.
	\citeauthor{nguyen_enhancing_2017} proposed a system consisting of a ToC page detection method and a link-based ToC reconstruction method to solve the ToC extraction problem.
	These works solved part of the document structure reconstruction problem, but the main body of one document, including content lines, figures, and tables, is not taken into consideration.
	
	The second one focuses on the overall structure reconstruction of document \cite{namboodiri_document_2007, wang_docstruct_2020, rausch_docparser_2021}. 
	\citeauthor{namboodiri_document_2007} proposed a framework of generic document structure understanding system in multi-steps. They split the processing procedure into two parts, including physical layout detection and logical structure recovering, with each part processed by digital image processing techniques and rule-based systems.
	\citeauthor{wang_docstruct_2020} focused on the form understanding task and considered the form structure as a tree-like hierarchy of text fragments. 
	They learned an asymmetric parameter matrix to predict the relationship between every text fragment pair, resulting in high computational complexity when handling documents with numerous text fragments.
	\citeauthor{rausch_docparser_2021} built a layout detection model based on Mask R-CNN \cite{MaskRCNN} and recovered structures of single document pages using a series of specific rules such as overlap ratio and reading flow. However, this method is difficult to apply to recover the structure of multi-page documents and lacks generalization due to the complicated heuristic rules introduced.

	\begin{figure*}[!htb]
		\centering
		\centering
		\includegraphics[width=1.0\linewidth]{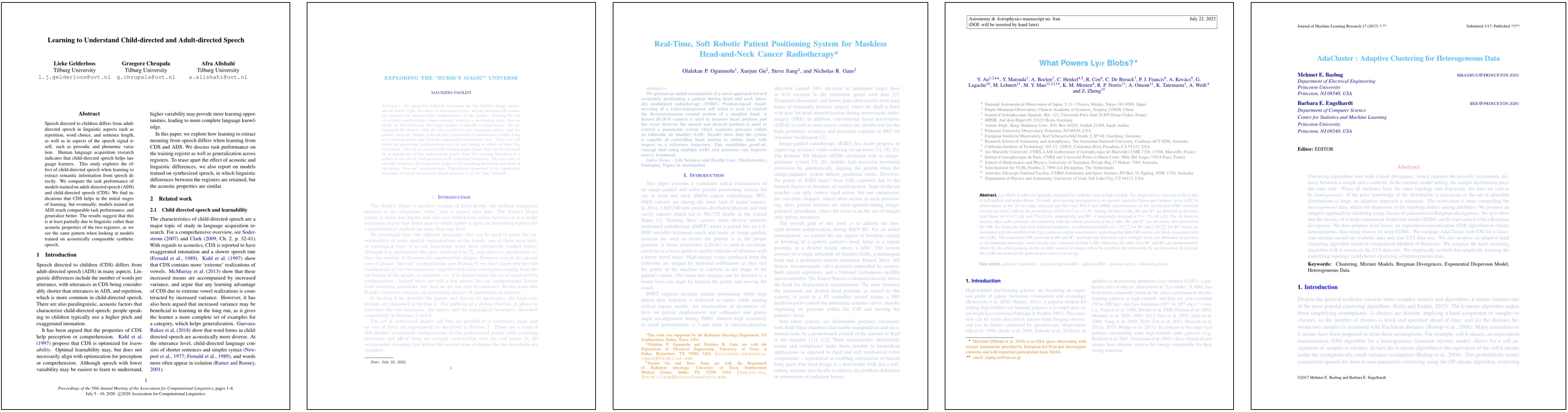}
		\centering
		\caption{Layout varieties of front document pages in the HRDoc dataset. The first front page is from the HRDoc-Simple dataset, with all the rest randomly selected from the HRDoc-Hard in a color version. }
		\label{fig:layout}
	\end{figure*}

	\section{The HRDoc Task and Dataset}
	In this section, we define the tasks introduced in HRDoc and then describe the data collection process.
	\subsection{Task Overview}
	We build HRDoc with line-level annotations and cross-page relations that support both NLP and CV research.
	HRDoc dataset aims to recover the semantic structure of the PDF document, which can be divided into three subtasks, including semantic unit classification, parent finding, and relation classification. 
	The detailed descriptions for these tasks are listed as followings:
	
	\begin{itemize}
		\item \textsc{Overall Task} (Document hierarchical reconstruction.) Given a multi-page document $D$, recover the semantic structure of this document.
		\item \textsc{SubTask 1}  (Semantic unit classification.) Given all semantic units $U$ with bounding boxes and texts within a document, classify each unit to its semantic label.
		\item \textsc{SubTask 2}  (Parent finding.) Given a semantic unit $u_i$ in a document, find its nearest parent unit $p_i$.
		\item \textsc{SubTask 3}  (Relation classification.) Classify the relation $r_i$ between each semantic unit and its parent unit.
	\end{itemize}

	\subsection{Dataset Collection}
	HRDoc dataset consists of two parts according to the layout varieties. 
	We name the first part HRDoc-Simple (HRDS), consisting of 1,000 documents with similar layouts, and the second part HRDoc-Hard (HRDH), consisting of 1,500 documents with varied layouts.
	
	To build the HRDS dataset, we download 1,000 conference papers in PDF format from ACL\footnote{https://aclanthology.org/} website, containing three top-tier conferences, including NAACL, EMNLP, and ACL in the natural language process field. 
	For the HRDH dataset, we downloaded 1,500 documents from arXiv\footnote{https://export.arxiv.org/} website, containing more than 30 varied layouts from 17 research areas. Moreover, thanks to the open-access policy of the arXiv website, we can download the LaTeX source code with PDF files simultaneously for further detecting different semantic regions.
	As shown in Figure \ref{fig:layout}, documents in HRDoc-Simple share a similar layout with the same ACL conference template. In contrast, documents in HRDoc-Hard have more varieties in page layout, containing single-column, two-column, and more complicated formats.
	
	Notably, all downloaded papers from the ACL website are licensed under CC BY-NC-SA 3.0\footnote{https://creativecommons.org/licenses/by-nc-sa/3.0/} or CC BY 4.0\footnote{https://creativecommons.org/licenses/by/4.0/}.
	For the arXiv website, we only download papers licensed under CC BY-NC-SA 4.0\footnote{https://creativecommons.org/licenses/by-nc-sa/4.0/} with compilable LaTeX source code. These licenses allow people to remix, transform, and build upon the material for non-commercial purposes.

	\subsection{Semantic Unit Parsing and Relation Definition}
	The unit classification task introduced in HRDoc is similar to that of PubLayNet  \cite{zhong_publaynet_2019} and DocBank \cite{li_docbank_2020}. This paper classifies all semantic units into 14 classes, including \{\textit{Title, Author, Mail, Affiliation, Section, First-Line, Para-Line, Equation, Table, Figure, Caption, Page-Footer, Page-Header, and Footnote}\}. 
	All class units are detected and labeled in text line level except for \textit{Equation, Table and Figure}, since they may contain multi-line texts.
	Each semantic unit comes after its corresponding parent unit when organized across every page in the human reading order. 
	The relation between one semantic unit and its parent unit consists of three types: \{\textit{Connect, Contain, Equality}\}. The definition of these relations is listed as follows:
	\begin{itemize}
		\item \textit{Connect} is used when two units are semantically connected. For example, the relation between continuous lines in a paragraph is called \textit{Connect}.
		\item \textit{Contain} is used when the child unit is one part of the parent unit. For example, the relation between a section line and its first subsection line is called \textit{Contain}.
		\item \textit{Equality} is used when two units are in the same hierarchical structure level. For example, the relation between different subsections under the same section is called \textit{Equality}.
	\end{itemize}

	\subsection{Ground-Truth Annotation}
	\label{pdf_parser}
	For the HRDS dataset, we design a rule-based PDF parser system to extract each text line in ACL papers using PDFPlumber\footnote{https://github.com/jsvine/pdfplumber}. 
	Owing to the fixed document layout of the HRDS dataset, we can roughly tag each text line to its labels using heuristic rules. For example, we can locate the first line of each paragraph by combining 1$)$ The distance between the current text line and previous text line; 2$)$ The indent before the beginning of the current text line; 3$)$ Whether the first character is upper-cased or lower-cased.
	
	For HRDH dataset, following \cite{li_docbank_2020}, we insert the `color' command with varied RGB parameters to different semantic code block and re-compile the LaTeX source code to get the colored PDF file as shown in Figure \ref{fig:layout}.
	After all PDF files are colored, we design another rule-based PDF parser system similar to that used in building the HRDS dataset. We can use both heuristic rules and color information of each character to label text lines roughly.
	To eliminate the influence of inserted `color' command such as line-shifting and font-size changing, we only keep papers whose text lines remain in their original position compared to those of downloaded PDF files.
	
	Since semantic units of certain classes (\textit{Equation, Table, and Figure}) contain more than one text line, 
	we use Cascade-RCNN \cite{CascadeRCNN} to locate their bounding boxes for both datasets.
	After all semantic units are located and roughly labeled, we ask the human annotators to recheck the labels and assign one parent unit with the proper relation for each semantic unit.
	
	\begin{figure}[!t]
		\centering
		\centering
		\includegraphics[width=0.9\linewidth]{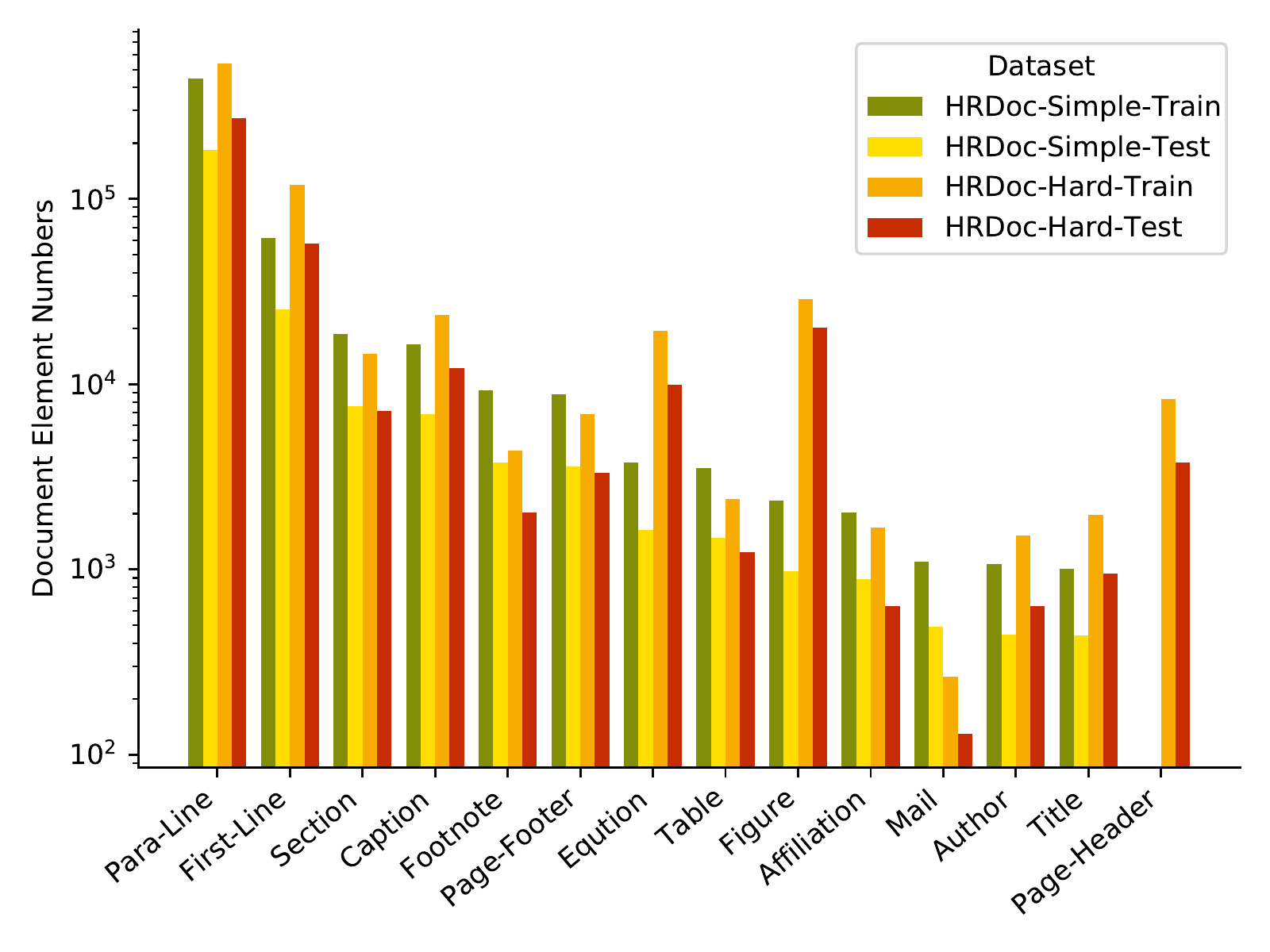}
		\caption{Statical results on numbers of different document semantic units of HRDoc dataset. }
		\label{fig:unit_num}
	\end{figure}

	\subsection{Statistic Analysis}
\begin{figure}[!t]
		\centering
		\centering
		\includegraphics[width=0.9\linewidth]{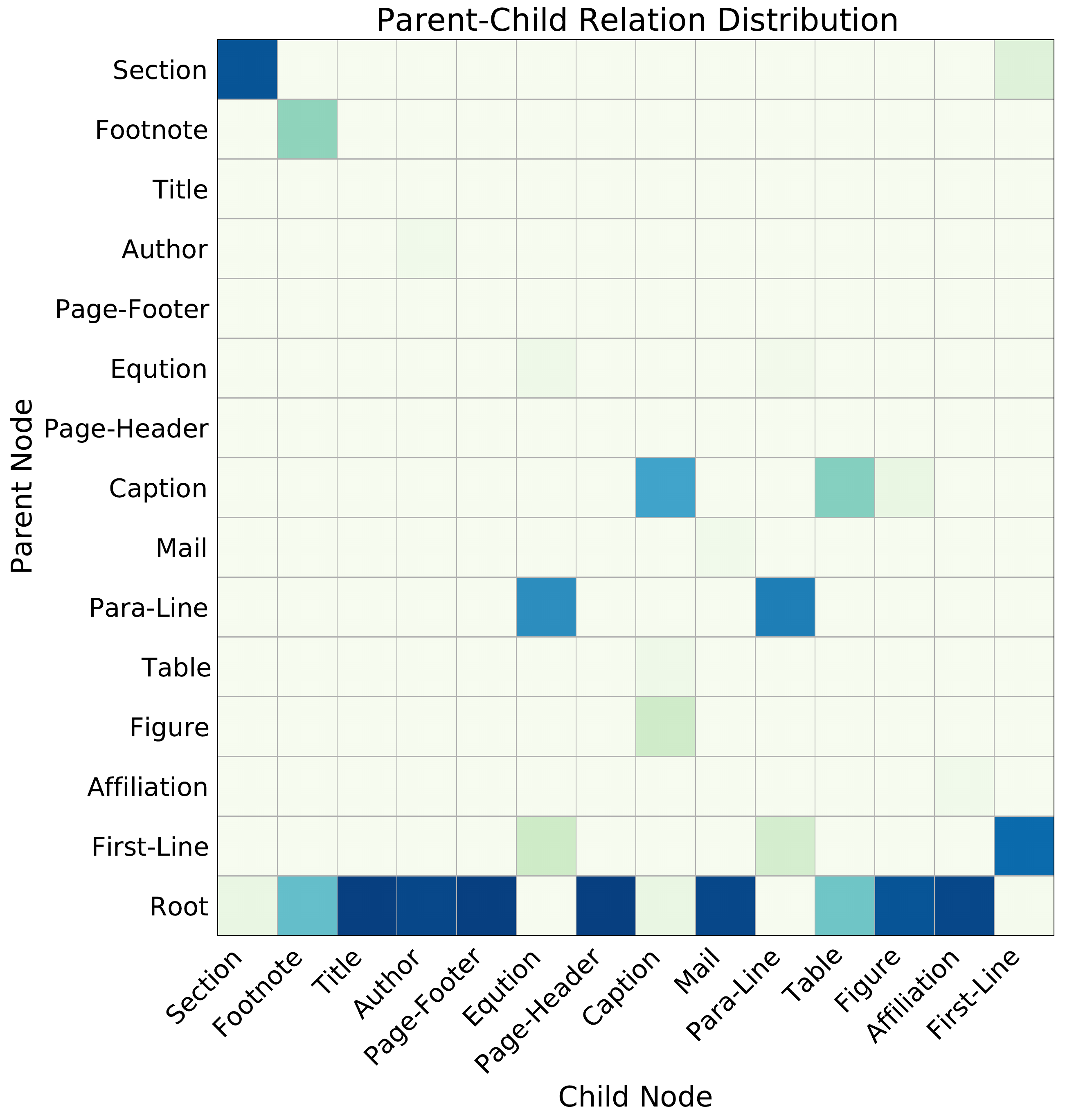}
		\caption{Statical results on parent ratio of different document semantic units of HRDoc dataset. } 
		\label{fig:parent_num}
	\end{figure}

	HRDoc contains 2,500 documents with nearly 2 million semantic units. Figure \ref{fig:unit_num} provides the statistics of semantic unit distribution over the train and test set of both HRDS and HRDH datasets.
	We can see that the \textit{Page-Header} entity is missing in HRDS dataset since the ACL template contains no header contents. Moreover, the distributions of all classes in HRDS and HRDH are pretty different.
	Figure \ref{fig:parent_num} provides the child-parent relation distribution over both HRDS and HRDH dataset. We can see that every semantic entity only has parents in certain entity types under the constraint of grammar rules.

	\begin{figure*}[!htb]
		\centering
		\centering
		\includegraphics[width=1.0\linewidth]{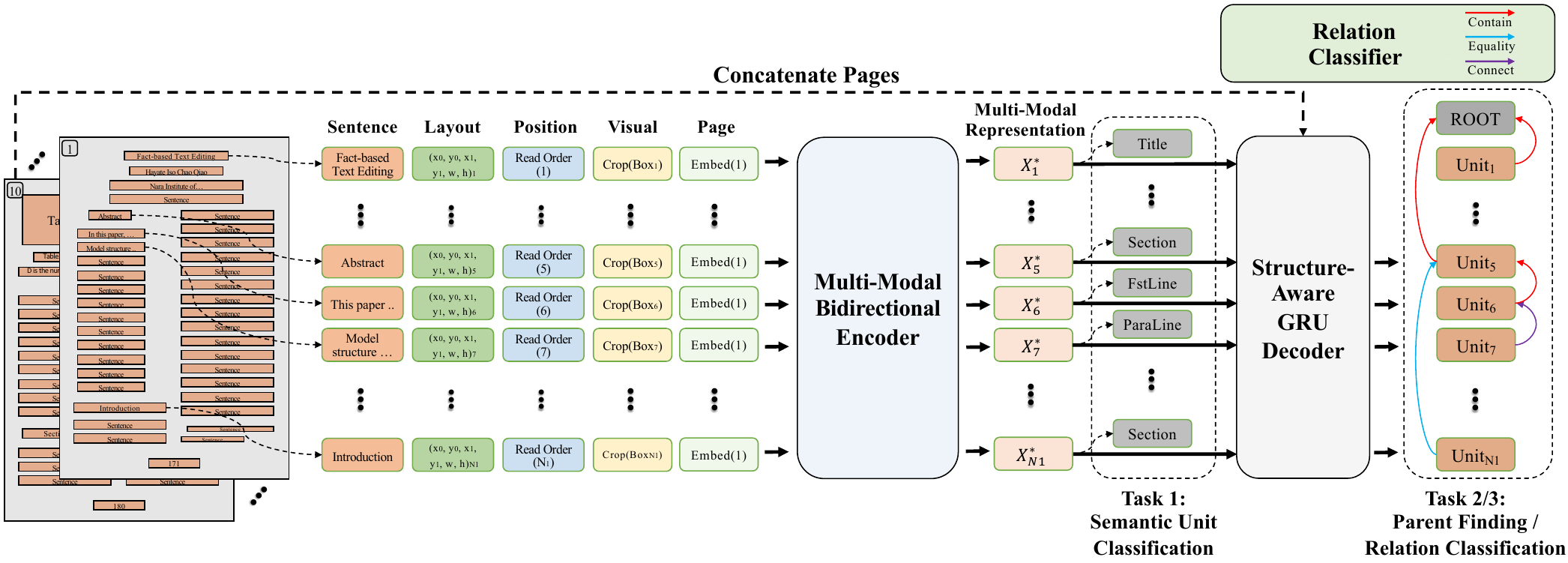}
		\centering
		\caption{Proposed document structure parsing system (DSPS).  }
		\label{fig:overall}
	\end{figure*}
	
	\section{Proposed Method}
	The proposed hierarchical document structure parsing system (DSPS) comprises a multi-modal bidirectional encoder, a structure-aware GRU decoder with soft-mask operation, and a relation classifier. In this section, we first introduce the main structure of the DSPS model and give a detailed description of sub-modules in the following subsections.
	
	As shown in Figure \ref{fig:overall}, 
	different types of embeddings for each semantic unit are calculated and fed into the encoder.
	After obtaining their multi-modal representations, these semantic units are further classified into different categories (\textsc{SubTask 1}). 
	Moreover, using a structure-aware GRU decoder with soft-mask operation, we can find the corresponding parent unit for each semantic unit (\textsc{SubTask 2}). 
	After obtaining parent-child pairs, we can classify the relation of each pair through the relation classifier module (\textsc{SubTask 3}). 
	The final structure of the whole document can be easily recovered through a post-processing procedure  (\textsc{Overall Task}).
	
	\subsection{Document Images and Semantic Units}
	
	To make use of visual information, we convert a $K$-page document $D$ into $K$ images $I = \{I_1, I_2, .., I_K\}$ for every page using PyMuPDF\footnote{https://github.com/pymupdf/PyMuPDF}. 
	As mentioned in Section \ref{pdf_parser}, we regard each text line, figure, table, and equation area in one document as the basic semantic unit.
	Each semantic unit $u_i$ for a $K$-page document $D$ can be identified by its text $s_i$, bounding box $b_i = (x_{\text{min}}, y_{\text{min}}, x_{\text{max}}, y_{\text{max}})_i$ and page number $p_i \in \{1,...,K\}$.
	
	The text $s_i$ is set to `Table' for \textit{Table} and `Figure' for \textit{Figure} units, while the other remains original texts extracted from PDF files using the bounding box $b_i$.
	Here $(x_{\text{min}}, y_{\text{min}})_i$ and $(x_{\text{max}}, y_{\text{max}})_i$ are the coordinates of the top-left and bottom-right corners of the $k$-th semantic unit's bounding box.
	We arrange each semantic unit in one document page in reading order, forming the input unit sequence $U = \{u_1, u_2, ... u_L\}$, where $L$ is the total unit number within the page with a max length limit of 512.
	
	\subsection{Multi-Modal Bidirectional Encoder}
	Similar as \cite{graphdoc}, the input embeddings of semantic units are consisted of a sentence embedding $x^{\text{text}}$, a layout embedding $x^{\text{layout}}$, a 1D positional embedding $x^{\text{pos}}$, a visual embedding $x^{\text{vis}}$ and a page embedding $x^{\text{page}}$. 
	
	In total, the embedding $x_i \in \mathbb{R}^H$ at the $i$-th position in the input sequence is given as:
	$$x_i = \text{LN}(x_i^{\text{text}}+ x_i^{\text{layout}} + x_i^{\text{pos}} + x_i^{\text{vis}} + x_i^{\text{page}}) $$
	Here $\text{LN}(\cdot)$ is the layer normalization operation \cite{LN}.
	\subsubsection{Sentence embedding.}
	We use Sentence-Bert \cite{SentenceBert} model as the semantic extractor.
	The sentence embedding can be calculated as follows:
	$$ x_i^{\text{text}} = \text{LinearProj}(\text{SentenceBert}(s_i)) $$
	Here the $\text{LinearProj}(\cdot)$ is a linear projection layer to transform the embedding dimension to $H$.
	\subsubsection{Layout embedding.}
	Following LayoutLMv2 \cite{layoutlmv2}, we introduce a location representation $x_i^{\text{layout}} \in \mathbb{R}^{H}$ that convey the layout information of $i$-th semantic token in the document page.
	$$x_i^{\text{layout}} = \text{Concat}(\text{Emb}(x_\text{min}, x_\text{max}, w)_i, \text{Emb}(y_\text{min}, y_\text{max}, h)_i)$$
	\subsubsection{Position embedding.}
	We use absolute position embedding here as 
	$$x_i^{\text{pos}} = \text{Emb1D}(i)$$
	\subsubsection{Visual embedding.}
	The visual clue of one semantic unit is also crucial since the font styles may differ between different semantic blocks.
	We use ResNet-50 \cite{ResNet} with FPN \cite{fpn} as the backbone of vision information extractor. 
	The local visual embedding can be obtained through RoIAlign \cite{MaskRCNN} according to $b_i$:
	$$x_i^{\text{vis}} = \text{RoIAlign}(\text{ResNet}(I_{p_i}), b_i)$$
	\subsubsection{Page embedding.}
	To distinguish semantic units of different pages, we use absolute position embedding here as 
	$$x_i^{\text{page}} = \text{EmbPage}(p_i)$$
	
	The multi-modal bidirectional encoder is based on Transformer \cite{Transformer} architecture and produces representations $x_i^{*}=\text{BiEncoder}(x_i)$ for each semantic unit.
	
	We can predict the label $\hat{C}_i \in \{1,2,\cdots,C\}$ for unit $u_i$ as follows:
	$$P_{\text{cls}_i} = \text{softmax}(\text{LinearProj}(x_i^{ *}))$$
	$$\hat{C}_i = \text{argmax}(P_{\text{cls}_i})$$
	where $P_{\text{cls}_i}$ means the classification probability for $i$-th unit over predefined $C$ classes.

	\subsection{Structure-Aware GRU Decoder}
	When reconstructing the hierarchical structure of one multi-page document, we often meet the cross-page parent-finding problem, which means a semantic unit may find its parent in previous pages. To address this problem, we first concatenate each semantic unit within a document in reading order and use a GRU \cite{GRU} network to capture the information exchange across pages.

	\begin{figure}[t!]
		\centering
		\centering
		\includegraphics[width=0.95\linewidth]{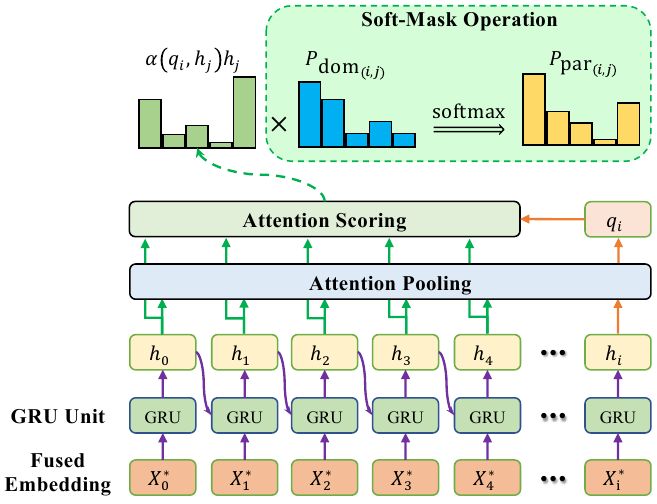}
		\caption{Proposed structure-aware GRU decoder. 
			}
		\label{fig:decoder}
	\end{figure}
	
	When decoding the parent unit, some semantic unit may directly link to the \textit{ROOT} unit, which represents the beginning of one document.
	The $x_{0}^{ *}$ referring to the multi-modal representation of the \textit{Root} unit is obtained using the average of all semantic units' representations:
	$$x_{0}^{ *} = \sum_{i=1}^{L}x_{i}^{ *}/L$$
	
	With multi-modal representations $x_i^{ *}$ as input, the GRU unit produces the hidden state $h_i$ of the current semantic unit:
	$$ h_i = \text{GRU}(x_{i-1}^{ *}, h_{i-1}) $$
	
	In decoding step $i$, we can calculate the weighted hidden state of $u_i$ using the attention pooling function:
	$$q_{i} = \sum_{j=0}^{i}\alpha(h_i, h_j)h_j$$
	where
	\begin{equation*}
		\begin{split}
			\alpha(h_i, h_j) & = \text{softmax}((\mathbf{W_q} h_i)^{T}\cdot \mathbf{W_k} h_j) \\
			& = \frac{\text{exp}((\mathbf{W_q} h_i)^{T}\cdot \mathbf{W_k} h_j)}{\sum_{j=0}^{i}\text{exp}((\mathbf{W_q} h_i)^{T}\cdot \mathbf{W_k} h_j)}
		\end{split}
	\end{equation*}
	
	\subsubsection{The soft-mask operation.} We introduce a soft-mask operation to adjust the attention distribution for better use of domain-specific knowledge. 
	More specifically, we calculate the child-parent distribution matrix $\mathbf{M_{cp}} \in \mathbb{R}^{(C+1)\times C}$ by counting different types of child-parent pairs as shown in Figure \ref{fig:parent_num}. The $i$-th column of $\mathbf{M_{cp}}$ means the probability distribution of class $i$ over $C$ predefined classes plus one \textit{ROOT} node. To make the distribution robust to unseen relation pairs, we apply additive smoothing to each column with the pseudo-count set to 5. The probability of semantic unit $u_j$ being the parent of $u_i$ can be calculated as:
	$$P_{\text{par}_{(i,j)}} = \text{softmax}(\alpha (q_i, h_j) h_jP_{\text{dom}_{(i,j)}})$$
	Here, $P_{\text{dom}_{(i,j)}}$ is defined as:
	$$P_{\text{dom}_{(i,j)}} = \widetilde{P}_{\text{cls}_j} \mathbf{M_{cp}} P^\mathrm{T}_{\text{cls}_i}$$
	where
	\begin{equation*}
		\widetilde{P}_{\text{cls}_j} =  
		\begin{cases}
			\text{Concat}(P_{\text{cls}_j}, 0) & j \in \{1,2,\cdots,i-1\};\\
			[\overbrace{0,0, ... , 0}^{C}, 1] & j =0.
		\end{cases}
	\end{equation*}
	
	We can calculate the parent node $\hat{P}_i \in \{0, 1,\cdots,i-1\}$ for $u_i$ as:
	$$\hat{P}_i = \text{argmax}(P_{\text{par}_{(i,j)}})$$
	\subsection{Relation Classifier}
	After obtaining the parent node $u_j$ for semantic unit $u_i$, we use a linear project function to classify the relation between each child-parent pair $(u_i, u_j)$ as:
	
	$$\hat{R}_{(i,j)} = \text{argmax}(P_{\text{rel}_{(i,j)}})$$
	$$P_{\text{rel}_{(i,j)}} = \text{softmax}(\text{LinearProj}(\text{Concat}(h_i,h_j)))$$
	where $P_{\text{rel}_{(i,j)}}$ means the classification probability for relation of child-parent pair $(u_i, u_j)$ over predefined $N$ relation types.
	
	\subsection{Multi-Task Learning}
	We take the three subtasks into consideration simultaneously.
	For \textsc{SubTask 1}, the loss function can be listed as:
	$$L_{\text{cls}} = \sum_{i=1}^{L}\text{FocalLoss}(C_{i}, P_{\text{cls}_{i}})/L$$
	For \textsc{SubTask 2}, the loss function can be listed as:
	$$L_{\text{par}} = \sum_{i=1}^{L}\text{CrossEntropy}(P_{i}, P_{\text{par}_{(i,j)}})/L$$
	For \textsc{SubTask 3}, the loss function can be listed as:
	$$L_{\text{rel}} = \sum_{i=1}^{L}\text{FocalLoss}(R_{i}, P_{\text{rel}_{(i,j)}})/L$$
	Here $C_{i}, P_{i}, R_{i}$ refer to the ground-truth label for the semantic class, parent position and the relation type for $u_i$ in one-hot representation.
	$\text{FocalLoss}(\cdot)$ \cite{focalloss} is also introduced to solve the class-imbalance problem.
	We calculate the weighted sum of these loss functions to form the final loss as:
	$$L_{\text{tol}} = L_{\text{cls}} + \alpha_1 L_{\text{par}} + \alpha_2 L_{\text{rel}}$$

	\begin{table*}[htp]
		\centering
		\scalebox{0.8} 
		{		
			\begin{tabular}{c ccccccc ccccccc cc}
				\toprule
				\multirow{2}{*}[-4pt]{Method} & \multicolumn{16}{c}{HRDoc-Simple \space \space \space F1(\%)}  \\ 
				\cmidrule(lr){2-17}
				& Title& Author & Mail& Affili&Sect&Fstl&Paral&Table&Fig&Cap&Equ&Foot&Head&Footn&Micro&Macro \\
				\midrule
				T1 & 78.83 & 72.74 & 64.54 & 70.13 & 91.35 & 87.53 & 89.70 & 89.30 & 73.87 & 64.87 & 83.87 & 87.50 & - & 79.32 & 88.30 & 80.85  \\
				T2 & 93.67 & 82.53 & 81.33 & 84.39 & 37.09 & 38.39 & 91.86 & 58.44 & 48.53 & 70.75 & 26.89 & 98.33 & - & 49.76  & 85.61 & 66.30 \\
				T3 & 98.98 & 96.47 & \textbf{98.95} & \textbf{97.42} & 97.30 & 93.27 & 98.72 & 94.42 & 95.72 & 93.36 & 96.02 & 99.89 & - & 87.11 & 97.74 & 95.97  \\
				T4 & \textbf{99.43} & \textbf{98.83} & 96.45 & 97.33 & \textbf{99.60} & \textbf{98.22} & \textbf{99.74} & \textbf{100.00} & \textbf{99.95} & \textbf{99.06} & \textbf{97.91} & \textbf{100.00} & - & \textbf{99.15} & \textbf{99.52} & \textbf{98.90}  \\
				\midrule
				\multirow{2}{*}[-4pt]{Method} & \multicolumn{16}{c}{HRDoc-Hard \space \space \space \space F1(\%)}  \\ 
				\cmidrule(lr){2-17}
				& Title& Author & Mail& Affili&Sect&Fstl&Paral&Table&Fig&Cap&Equ&Foot&Head&Footn&Micro&Macro \\
				\midrule
				T1 & 81.50 & 49.77 & 33.39 & 49.34 & 75.92 & 64.96 & 77.86 & 69.96 & 72.22 & 43.72 & 68.84 & 70.91 & 71.00 & 52.67 & 73.37 & 64.94  \\
				T2 & 82.40 & 48.40 & 18.43 & 61.33 & 33.66 & 45.37 & 87.99 & 21.89 & 70.28 & 61.54 & 48.32 & 73.69 & 75.71 & 6.79& 79.25 & 52.56   \\
				T3& 95.85 & 89.92 & \textbf{91.68} & \textbf{91.75} & 94.26 & 88.68 & 96.77 & 76.96 & 91.67 & 91.99 & 93.94 & 94.68 & 92.65 & 62.61 & 94.68 & 89.53  \\
				T4& \textbf{97.71} &\textbf{93.93} & 85.49 & 90.95 & \textbf{96.06} & \textbf{91.24} & \textbf{97.96} & \textbf{100.0} & \textbf{100.0} & \textbf{97.32} & \textbf{97.92} & \textbf{98.54} & \textbf{97.83} & \textbf{88.84} & \textbf{96.74} & \textbf{95.27}  \\
				\bottomrule
			\end{tabular}
		}
		\caption{Comparison results of different baseline models in semantic unit classification task. F1 means F1-score. T1 refers to Cascade-RCNN, T2 refers to ResNet+RoIAlign, T3 refers to Sentence-Bert, T4 refers to DSPS Encoder.}
		\label{tab:classify_compare}
	\end{table*}
	
	\begin{table*}[htp]
		\centering
		\scalebox{0.8} 
		{
			\begin{tabular}{l cccccccc}
				\toprule
				\multirow{2}{*}[-4pt]{Method} &\multirow{2}{*}[-4pt]{Semantic} &\multirow{2}{*}[-4pt]{Vision} & \multirow{2}{*}[-4pt]{Soft-Mask} & \multirow{2}{*}[-4pt]{Level}  & \multicolumn{2}{c}{HRDoc-Simple}   & \multicolumn{2}{c}{HRDoc-Hard}  \\ 
				\cmidrule(lr){6-7}\cmidrule(lr){8-9}
				&&&&&Micro-STEDS&Macro-STEDS&Micro-STEDS&Macro-STEDS \\ 
				\midrule
				DocParser & - & - & - & Page & 0.2361 & 0.2506 & 0.1873 & 0.2015 \\
				DSPS & \xmark & \cmark & \xmark & Document & 0.6149 & 0.6284 & 0.5145 & 0.5393 \\
				DSPS & \cmark & \xmark & \xmark & Document & 0.6636 & 0.6690 & 0.5766 & 0.5817 \\
				DSPS & \cmark & \cmark & \xmark & Document & 0.6830 & 0.6888 & 0.5811 & 0.5968 \\
				DSPS & \cmark & \cmark & \cmark & Document & \textbf{0.8143} & \textbf{0.8174} & \textbf{0.6903} & \textbf{0.6971} \\
				DSPS & \cmark & \cmark & \cmark & Page & 0.6482 & 0.6562 & 0.6080 & 0.6243 \\
				\bottomrule
			\end{tabular}
		}
		\caption{Comparison results of different models in hierarchical document reconstruction task.}
		\label{tab:e2e_compare}
	\end{table*}
	%
	%

	\section{Experiments}
	
	\subsection{Compared Methods}
	\subsubsection{Semantic unit classification.}
	The \textsc{SubTask 1} of semantic unit classification can be viewed as both CV and NLP tasks.
	To this end, we use a state-of-the-art detection model Cascade-RCNN \cite{CascadeRCNN} to detect the position of each semantic unit.
	With bounding boxes and texts of each semantic unit given, we use the features outputted by ResNet-50 with RoIAlign and Sentence-Bert to classify each semantic unit.
	The proposed multi-modal bidirectional encoder is also used to classify each semantic unit.


	\subsubsection{Hierarchical document reconstruction.}
	To evaluate the performance of DocParser \cite{rausch_docparser_2021} system in \textsc{Overall Task}, we do the inference stage of DocParser in HRDoc dataset by excluding the table structure and retrieving semantic units for each detected objects with an overlap ratio higher than 0.7.
	We also test the performance of the DSPS model with different settings.
	
	\subsection{Evaluation Metrics}
	\subsubsection{Semantic unit classification.}
	We use the standard F1 score on each category to evaluate the semantic unit classification task performance.
	For the output of the detection model, we only preserve those predicted bounding boxes with a confidence coefficient higher than 0.5. An output box is regarded as correctly predicted only when it has an overlap ratio higher than 0.65 with a ground-truth box in the same label.
	
	\subsubsection{Hierarchical document reconstruction.}
	For a given document $D$, we can get its content structure $T_D$ and predicted structure $\hat{T}_D$ in a tree-like format. 
	Inspired by \cite{TEDS}, 
	we proposed the Semantic-TEDS (Tree-Edit-Distance-Score) metric to evaluate the system performance on the document reconstruction task. 
	$$\text{STEDS}(T_D, \hat{T}_D) = 1-\frac{\text{EditDist}(T_D, \hat{T}_D)}{\text{max}(|T_D|, |\hat{T}_D|))}$$
	Here $|T|$ refers to the number of nodes in a tree structure $T$. Notice that we only regard one node in $T_D$ and $\hat{T}_D$ as completely matched when the label and text are the same.


	\subsection{Evaluation Results}
	\subsubsection{Semantic unit classification.}
	The evaluation results of compared methods are shown in Table \ref{tab:classify_compare}.
	We can see that the proposed multi-modal bidirectional encoder has the best classification performance in most categories. 
	Sentence-Bert outperforms the DSPS encoder in specific categories, including \textit{Mail} and \textit{Affili}, which indicates that the visual clue may be harmful when facing visually similar units.
	Due to the complexity of document layouts in the HRDH dataset, all models show a decrease when dealing with the HRDH dataset compared to the HRDS dataset.
	
	\subsubsection{Hierarchical document reconstruction.}
	We conducted experiments on DocParser and the proposed DSPS model with different settings.
	As shown in Table \ref{tab:e2e_compare}, DSPS model outperform DocParser by a large margin. 
	We can see that both semantic and visual modalities are crucial for the DSPS model. Moreover, the soft-mask operation introduced in the decoder also brings considerable improvement.
	In the page-level setting for DSPS, we force the decoder to find the parent unit within the same document page for each semantic unit. 
	The system performance drops significantly in the page-level setting, indicating that the document reconstruction task should be conducted in a cross-page manner.
	
	\section{Conclusion}
	This paper introduced the hierarchical reconstruction of document structures as a novel vision and language task. 
	To facilitate research aimed at document reconstruction, we proposed a new dataset named HRDoc, which contains 2,500 line-level annotated documents with nearly 2 million semantic units. 
	The proposed DSPS model is based on an encoder-decoder structure that takes multi-modal information as input. With both semantic and vision features considered, DSPS achieves the best classification performance surpassing other methods by a large margin.
	In the overall task, we proposed a structure-aware GRU decoder with domain-specific knowledge introduced.
	By training in an end-to-end manner, the DSPS model achieves a considerable improvement over the baseline model.
	
	\bibliography{aaai23}
	
\end{document}